\documentclass[11pt,a4paper]{article}
\usepackage[T1]{fontenc}
\usepackage{lmodern}
\usepackage{mathptmx}
\usepackage[margin=25mm]{geometry}
\usepackage{amsmath,amssymb}
\usepackage{graphicx}
\usepackage{adjustbox}
\usepackage{booktabs}
\usepackage{multirow}
\usepackage{array}
\usepackage{xcolor}
\usepackage[labelfont=bf,labelsep=space]{caption}
\usepackage[numbers,sort&compress]{natbib}
\usepackage{authblk}
\usepackage{pdfpages}
\usepackage[hidelinks]{hyperref}

\newlength{\figwidth}
\title{Do Waste Classifiers Recognise Items They Have Never Seen? Object-Disjoint Evaluation on Household and Medical Waste}
\author[1]{Nusrat Jahan}
\author[1]{Mst. Eshrat Jahan Esha}
\author[1]{Masuma Sadia}
\author[1]{Saad Ahmed\thanks{Corresponding author: saadahmed@baust.edu.bd}}
\author[1]{Md Zahim Hassan}
\author[1]{S M Jahangir Alam}
\affil[1]{Department of Computer Science and Engineering, Bangladesh Army University of Science and Technology (BAUST), Saidpur Cantonment, Saidpur 5310, Nilphamari, Bangladesh}
\date{Preprint. Submitted to the Journal of Material Cycles and Waste Management.}

\begin{document}
\maketitle

\begin{abstract}
Automatic waste classification supports source separation, which enables recycling and keeps hazardous medical items out of household waste. Waste classification studies often report near-perfect accuracy on random image splits, which put photographs of one item in both training and test sets. To measure how well classifiers recognise items they have never seen, we built BanglaWaste-23, with 4,160 images of 172 objects in 23 household and medical waste classes from Bangladesh. Every image also records which physical item it shows. We evaluated 11 ImageNet-pretrained models and two models trained from scratch (a compact custom network and MobileNetV3-L) using object-disjoint five-fold cross-validation and a random split. On unseen objects, the best pretrained model reached a macro-averaged F1 score of 98.8\%, and six models were statistically tied. The random split overstated from-scratch accuracy by 15 to 19 percentage points. The from-scratch models also missed 11 to 13\% of medical items, against under 1\% for pretrained models. On a second dataset from Dhaka, pretrained accuracy fell by 15 to 22 points. Classifiers should therefore be evaluated on unseen objects, with random-split results read as an upper bound. The data are on Zenodo and the code at \url{https://github.com/saadbaust/BanglaWaste-23}.

\end{abstract}

\section{Introduction}\label{sec:intro}
Municipal solid waste is growing quickly. The world produced about 2.1 billion tonnes of it in 2020, and this amount is
expected to reach 3.8 billion tonnes by 2050 \citep{unep2024gwmo}. Much of this waste could be recycled or composted if it were
separated where it is produced. Separation at the source is therefore a central step in waste management. Automatic recognition
of discarded items could support this step.

Medical waste makes separation more urgent. About 85\% of the waste from health-care activities is comparable to household waste. The remaining 15\% is hazardous because it may be infectious, chemical or radioactive \citep{who2024healthcarewaste}. This hazardous part must be kept apart from household waste. In Bangladesh, national plans call for waste segregation from households to landfills. These plans, however, are not yet
fully implemented \citep{jerin2022dhaka}. In one district hospital, infectious waste that had been separated at the source was
later mixed with other waste and handed over for municipal dumping \citep{sujon2022medical}. In a hospital in Rajshahi, only a third of the waste bins were fully colour-coded \citep{tabassum2026rajshahi}. Hospital items and household items
can therefore end up in the same waste stream.

Image-based classification is a promising way to automate sorting. Many studies have trained deep learning models on waste images, and they often report
very high accuracies \citep{rahman2024bdwaste,gurcan2026automated,madhavi2025swinconvnext,alkilinc2026ensemble,goyal2025vgd6net,utku2026trash}.

These accuracies, however, depend on how the test images were chosen. Waste datasets usually contain several photographs of each
physical item, taken from different angles or on different backgrounds. Most waste studies divide such datasets at random, image
by image (Section~\ref{sec:rw_leakage}). Views of the same item then appear in both the training and the test set. A model can
therefore succeed by recognising an item that it has already seen.  In other fields,
leaky splits of this kind are known to inflate results \citep{kapoor2023leakage,yagis2021leakage}. Two further gaps limit
current work. Only one public dataset that we found contains medical items collected in a hospital \citep{zhou2022medical}. In
addition, models are rarely tested on images from another region.

In this paper, we address these gaps with a new dataset and a careful benchmark. Our aim is to measure how well waste classifiers recognise physical items that they have never seen, and how much a random image-level split changes this picture. Our contributions are as follows. First, we release BanglaWaste-23, a dataset of 4,160 images of 172 physical objects in 23
household and medical waste classes, with an object ID and capture metadata for every image. Second, we define an
object-disjoint five-fold benchmark with released split files and evaluate 11 pretrained models and two models trained from
scratch under one tuned recipe. Third, we show how much a random image-level split overstates accuracy on unseen objects, and
why this effect differs between pretrained models and models trained from scratch. Fourth, we analyse medical-waste errors,
per-object difficulty, image-quality shortcuts, model attention and transfer to a second Bangladeshi dataset from another region.

The rest of the paper is organised as follows. Section~\ref{sec:related} reviews related work. Section~\ref{sec:dataset}
describes the dataset, and Section~\ref{sec:methods} the benchmark design. Section~\ref{sec:results} reports the results,
Section~\ref{sec:discussion} discusses them, and Section~\ref{sec:conclusion} concludes the paper. Figures and tables numbered with an S (for example, Fig.~S1 and Table~S1) are given in Online Resource 1.

\section{Related work}\label{sec:related}

\subsection{Waste image datasets}\label{sec:rw_datasets}
Public waste image datasets differ mainly in where and how the images were taken (Table~\ref{tab:datasets}). TrashNet
contains 2,527 images of six material classes, photographed one item at a time on a white board \citep{yang2016trashnet}.
RealWaste contains 4,808 images of nine classes taken at a landfill in Australia \citep{single2023realwaste}. TACO collects
crowdsourced photographs of litter in the wild and annotates them for detection \citep{proenca2020taco}. ZeroWaste records
cluttered items on the conveyor belt of a recycling facility \citep{bashkirova2022zerowaste}. Other datasets are built from web images or merged from earlier sets \citep{kumsetty2022trashbox,tanvir2025trashnext,kunwar2026garbage}. TrashBox, for example, has 17,785 web images, and TrashNeXt merges public sets into 23,625 images.  In such collections, it is hard to tell whether two images show the same item.

Medical waste appears in only a few of these datasets. \citet{zhou2022medical} collected 3,480 images of eight medical classes
in a hospital in China. The Medical Waste 4.0 dataset contains photographs of new, unused medical items taken in two rooms
\citep{bruno2023medical}. In TrashBox, ConvoWaste and TrashNeXt, the medical images come from the web or from merged public
sets \citep{kumsetty2022trashbox,nafiz2023convowaste,tanvir2025trashnext}. Bangladesh has one earlier waste image dataset.
BDWaste contains 2,497 images of 20 classes collected from roadsides and dustbins in Dhaka \citep{rahman2024bdwaste}. It
includes gloves, masks and empty medicine packets. It contains no items collected at a hospital. Finally, none of the datasets
in Table~\ref{tab:datasets} releases an object identity for each image.

\begin{table*}[!htbp]
\centering
\scriptsize
\caption{Public waste image datasets compared with BanglaWaste-23 (evaluation split as used in the original paper).}
\label{tab:datasets}
\begin{adjustbox}{max width=\textwidth}
\begin{tabular}{@{}lrrrllll@{}}
\toprule
Dataset & Year & Images & Classes & Source & Medical waste & Object IDs & Evaluation split \\
\midrule
TrashNet \citep{yang2016trashnet} & 2016 & 2,527 & 6 & Real, staged on white board & No & No & Not stated \\
TACO \citep{proenca2020taco} & 2020 & 1,500 & 60 & Real, crowdsourced litter & No & No & Random, image-level \\
TrashBox \citep{kumsetty2022trashbox} & 2022 & 17,785 & 7 & Web-scraped & Yes (web) & No & Not stated \\
ZeroWaste-f \citep{bashkirova2022zerowaste} & 2022 & 4,661 & 4 & Real, recycling conveyor & No & No & Not stated \\
Zhou et al. \citep{zhou2022medical} & 2022 & 3,480 & 8 & Real, hospital (China) & Yes (all) & No & Random, image-level \\
RealWaste \citep{single2023realwaste} & 2023 & 4,808 & 9 & Real, landfill & No & No & Random, image-level \\
Medical Waste 4.0 \citep{bruno2023medical} & 2023 & >1,400 & 7 & Real, new unused items & Yes (all) & No & Different-day test set \\
BDWaste \citep{rahman2024bdwaste} & 2024 & 2,497 & 20 & Real, street (Dhaka) & Street-collected & No & Random, image-level \\
TrashNeXt \citep{tanvir2025trashnext} & 2025 & 23,625 & 9 & Merged public sets & Yes (merged) & No & Random, image-level \\
Garbage Dataset \citep{kunwar2026garbage} & 2026 & 12,259 & 10 & App, web and community & No & No & Random, image-level \\
\textbf{BanglaWaste-23 (ours)} & \textbf{2026} & \textbf{4,160} & \textbf{23} & \textbf{Real, households and hospital} & \textbf{Yes (4 classes)} & \textbf{Yes (172 objects)} & \textbf{Object-disjoint 5-fold} \\
\bottomrule
\end{tabular}
\end{adjustbox}
\end{table*}

\subsection{Deep learning for waste classification}\label{sec:rw_models}
Most recent waste classification studies fine-tune convolutional neural networks (CNNs) or transformers that were
pretrained on ImageNet \citep{russakovsky2015imagenet}.
\citet{rahman2024bdwaste} trained MobileNetV2 and InceptionV3 on BDWaste and reported accuracies of up to 99.70\%.
\citet{gurcan2026automated} compared 14 pretrained CNNs on a 12-class Kaggle dataset, and ConvNeXt was the best model with
96.95\%. \citet{alkilinc2026ensemble} combined several CNNs in ensembles and reached 95.8 to 99.1\% on four public datasets.
\citet{goyal2025vgd6net} designed a custom CNN for six waste classes and reported an accuracy of 98.54\%.
Transformer and hybrid designs have also been used, such as a vision transformer on TrashNet \citep{huang2021vit}, a Swin and
ConvNeXt hybrid \citep{madhavi2025swinconvnext} and CNN and vision transformer hybrids for nine waste types \citep{utku2026trash}. In parallel, several studies propose compact custom CNNs, sometimes trained
with knowledge distillation \citep{zheng2023focusrcnet}. Lightweight CNNs have also been paired with explanation methods such as local interpretable model-agnostic explanations (LIME) \citep{ahmed2026lightweight}. Other studies detect and locate several waste items in one image. \citet{chen2021transfer} used transfer learning for this task on a self-built dataset. \citet{xiao2024fusion} created training images by fusing waste items onto complex backgrounds. For medical waste, \citet{zhou2022medical} reached 97.2\% with a
fine-tuned ResNeXt-50.

The way these results are reported makes them hard to compare and to carry over to practice. Most studies give a single
accuracy from one split, without a confidence interval. Only two of the studies that we reviewed ran significance tests. Some
fine-tune only the classifier layer and keep the pretrained backbone frozen \citep{rahman2024bdwaste,gurcan2026automated}.
Measured speed on a phone or an embedded device is reported only rarely \citep{huang2021vit,chaconalbero2025sensors}. When
images from several real sources were combined, classification accuracy reached only 74.6\% \citep{majchrowska2022waste}. A
review of computer vision for waste sorting also noted that many studies use simplified settings or artificially collected
data \citep{lu2022review}.

\subsection{Data leakage and shortcut learning}\label{sec:rw_leakage}
Data leakage occurs when information about the test data is available during training. \citet{kapoor2023leakage} documented
leakage in machine-learning studies across many scientific fields and grouped its causes into a taxonomy. In brain imaging,
splitting the image slices of one patient between training and test sets inflated classification accuracy compared with a
split by patient \citep{yagis2021leakage}. \citet{varoquaux2022failures} describe such evaluation failures in medical imaging and
recommend practices that avoid them. A related problem is shortcut learning, in which a model relies on a simple cue, such as
the background or the image quality, instead of the object itself \citep{geirhos2020shortcut}. Visual explanation methods such as gradient-weighted class activation mapping (Grad-CAM) help to check where a model
looks \citep{selvaraju2017gradcam}.

Waste classification has not yet adopted this kind of evaluation. Of the 30 waste image studies that we reviewed, 20 evaluated
on image-level splits without any grouping. Another 9 did not describe the split. Only one added a test set recorded on a different
day \citep{bruno2023medical}. None split the data by physical object or reported how many items were shared between the
training and test sets. We also found no study that trained on one dataset and tested on a dataset from another region. BanglaWaste-23 and its object IDs are designed to fill this gap.

\section{The BanglaWaste-23 dataset}\label{sec:dataset}

\subsection{Collection}\label{sec:collection}
We collected waste items that are commonly discarded in northern Bangladesh from households, student hostels, a campus
canteen and streets in Saidpur, Nilphamari District. The sites were the
Kazipara Panir Tank area, Cant Bazar and Bir Protik Taramon Bibi Hall. The last is a student residence of the Bangladesh Army
University of Science and Technology (BAUST). Medical waste items were collected at Hasan
General Hospital, Thakurgaon. The Managing Director of the hospital gave permission for this collection. Team members wore
gloves and face masks, and the items were returned to the hospital after photography. The
workflow from collection to release is shown in Fig.~S1 in Online Resource 1.

Each item was photographed several times with smartphones running the Open Camera application. A Samsung Galaxy M31 took
1,899 images and a Samsung Galaxy M32 took 1,992 images. The remaining 269 images were taken with a Xiaomi Redmi Note 11 and a vivo Y20i. The items were photographed from different
distances and angles, and many were rearranged or moved to another surface between shots.

The images are released at the resolution of the camera. Most of them (87.8\%) are $3456 \times 3456$ or $2992 \times 2992$
pixels, and every image is at least 513 pixels on its shorter side. Users can therefore choose their own input size,
unlike in datasets released after resizing, such as TrashNet \citep{yang2016trashnet} and RealWaste
\citep{single2023realwaste}.

\subsection{Classes and groups}\label{sec:classes}
\begin{figure}[!htbp]
\centering
\includegraphics[width=\figwidth]{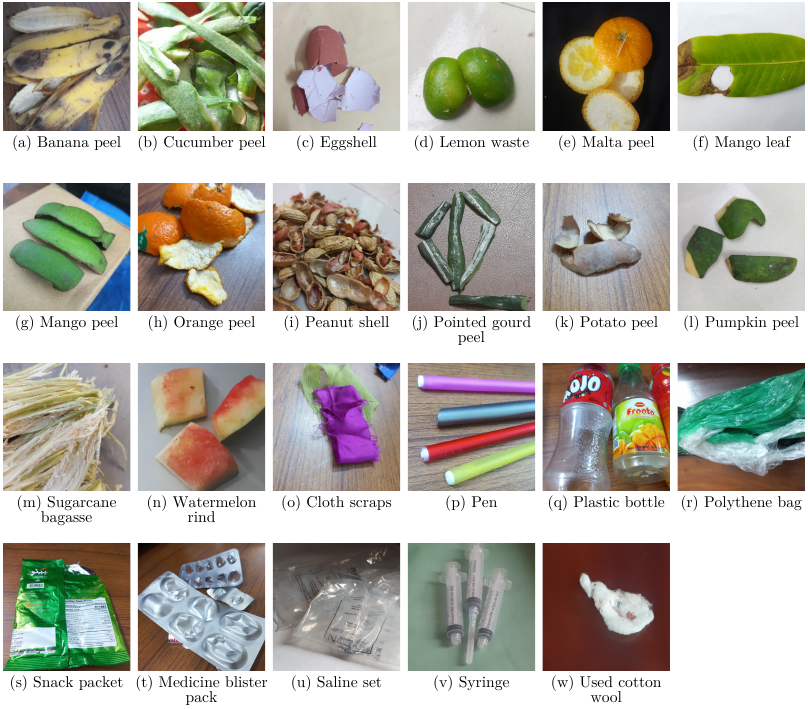}
\caption{One example image of each of the 23 classes of BanglaWaste-23: (a)--(n) food and plant waste, (o)--(s) household non-food
waste, and (t)--(w) medical waste (medicine blister pack, saline infusion set, syringe and used cotton wool).}
\label{fig:samples}
\end{figure}

BanglaWaste-23 contains 23 classes in three groups that matter for separation at the source. The food and plant group has 14 classes (banana peel, cucumber peel, eggshell, lemon waste, malta peel, mango leaf, mango peel, orange peel, peanut shell, pointed gourd peel, potato peel, pumpkin peel, sugarcane bagasse and watermelon rind) and 2,412 images (58.0\%). The household non-food group has five classes (cloth scraps, pen, plastic bottle, polythene bag and snack packet) and 967 images (23.2\%). The medical group has four classes (medicine blister pack, saline infusion set, syringe and used cotton wool) and 781 images (18.8\%). One example
image of each class is given in Fig.~\ref{fig:samples}, and the number of images per class in Fig.~S2.

The classes are well balanced, with 141 to 211 images each (mean 180.9). The largest class is only 1.5 times the size of the
smallest. Twelve of the classes do not appear in the BDWaste dataset from Dhaka \citep{rahman2024bdwaste}. These are cucumber peel, mango leaf, orange peel, peanut shell, pointed gourd peel, pumpkin peel, watermelon rind, cloth scraps, pen, saline infusion set, syringe and used cotton wool. Several food classes are
fine-grained. Potato, mango, cucumber and pointed gourd peels, for example, look alike in colour and texture.

We define an object as one physical item, or one set of items of the same class photographed together. One pile of peanut
shells, for example, is one object. The dataset contains 172 objects, with 5 to 10 objects per class. 

\subsection{Screening and labelling}\label{sec:screening}
Images were labelled by placing them in one folder per class. A nurse at Hasan General Hospital verified the four medical classes.

We then screened all 4,981 photographs. First, we removed images that were blurred, too dark, taken by mistake or showing
the wrong item. Next, we compared the MD5 checksums of all
files, which are short fingerprints of the file content. Six pairs of identical files were found, one in the bottle class and
five in the malta peel class. Both copies of each pair were removed, which is 12 files. Finally, we removed four peanut shell images whose shorter side was 510 or 511 pixels. These
steps removed 821 photographs (16.5\%) and left 4,160 images.

The released images were not resized, recompressed, colour-corrected or augmented. The only edit is the cropping of 69 images
(56 peanut shell and 13 bottle images), which are marked in the metadata. In addition, 2,228 images carry an exchangeable image file
format (EXIF) tag that tells software how to rotate the image. We applied this tag before training
(Section~\ref{sec:training}).

\subsection{Object identities and metadata}\label{sec:objects}
Because each item was photographed many times, we grouped the screened images into objects. First, photographs taken in
sequence within each class were grouped automatically by visual similarity. Similarity was measured with features from an
ImageNet-pretrained MobileNetV2 network \citep{sandler2018mobilenetv2} and with colour histograms of the image border. Next,
the team reviewed the proposed groups on contact sheets, which are pages of small thumbnails. Groups that showed the same items
in a different arrangement or on another background were merged, which gave 172 objects.

For each object, we also recorded the setting, the lighting and the background by visual inspection. Almost all objects (170 of
172) were photographed indoors, and most of them (144) in natural daylight. Sixty-three objects were photographed on more than
one background.

Object IDs make a fair test on new items possible. Without them, a random split puts views of the same object in both the
training and the test set. With them, whole objects can be kept out of training (Section~\ref{sec:protocol}), and errors
can be analysed object by object (Section~\ref{sec:res_hard}). None of the public waste datasets in
Table~\ref{tab:datasets} releases such IDs.

\section{Benchmark design}\label{sec:methods}

\subsection{Evaluation protocol}\label{sec:protocol}
\begin{figure}[!htbp]
\centering
\includegraphics[width=\figwidth]{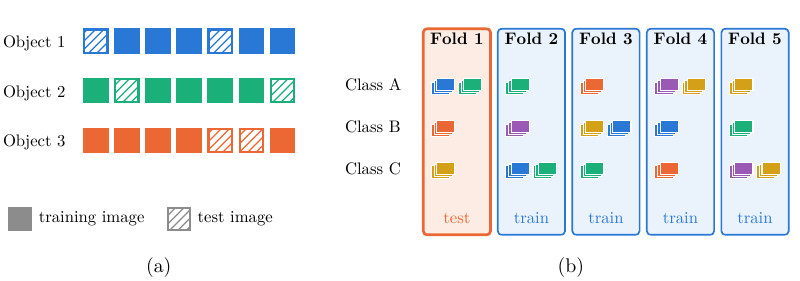}
\caption{Random image-level splitting versus the object-disjoint protocol used in this study. Each square is one image, and squares of
the same colour show the same physical object. (a) A random split places images of the same object in both the training and the test set. In BanglaWaste-23, 155 of the 172 objects end up on both sides, so the model is tested on objects it has already seen.
(b) In object-disjoint five-fold cross-validation, whole objects are assigned to folds, every class is present in every fold,
and each fold is used once as the test set (shown for fold 1). Learning rate and epochs were tuned on one object per class held
out from the training folds, never on the test fold, and the out-of-fold predictions of all 4,160 images were pooled.}
\label{fig:protocol}
\end{figure}

The main evaluation tests every model on objects that it has never seen. We used five-fold cross-validation in which whole
objects, not single images, were assigned to folds (Fig.~\ref{fig:protocol}b). Within each class,
objects were sorted from largest to smallest, and the first five were placed one per fold. Each further object went to the
fold with the fewest images of that class (random seed 42). As a result, every class appears in every fold, and no object
appears in two folds. The folds contain 740 to 936 images from 32 to 38 objects (Fig.~S3 in Online Resource 1). A standard stratified group split
could not achieve this, because several classes have only five objects of unequal size. Each fold served once as the test set,
and the other four folds were used for training.

Learning rates and the number of epochs were chosen without using any test fold. For this purpose, one object per class was
held out from the training folds of the first round, in which fold 1 is the test fold. This tuning split has 2,689 training
images and 623 validation images from 23 held-out objects.

For comparison, we also built the kind of split that most waste studies use (Section~\ref{sec:rw_leakage}). Images were
assigned at random to training, validation and test sets (70\%, 15\% and 15\%, stratified by class). In this split, 155 of
the 172 objects have images in both the training and the test set (Fig.~\ref{fig:protocol}a). We used it only to measure how
a random image-level split changes the results (Section~\ref{sec:res_leakage}).

\subsection{Models}\label{sec:models}
We benchmarked 11 models pretrained on the ImageNet-1k dataset \citep{russakovsky2015imagenet}. The mobile CNNs are
MobileNetV2 \citep{sandler2018mobilenetv2}, MobileNetV3-L \citep{howard2019mobilenetv3}, MobileNetV4-S
\citep{qin2024mobilenetv4}, EfficientNet-B0 \citep{tan2019efficientnet} and GhostNetV2 \citep{tang2022ghostnetv2}. The larger
CNNs are ResNet-50 \citep{he2016resnet}, DenseNet-121 \citep{huang2017densenet} and ConvNeXt-T \citep{liu2022convnext}.
DeiT-S is a vision transformer \citep{touvron2021deit}. MobileViTv2 \citep{mehta2023mobilevitv2} and EfficientFormerV2-S0
\citep{li2023efficientformerv2} combine convolution with attention. The models range from 2.25 to 27.84 million parameters
and from 166 to 4,599 million multiply-accumulate operations (MACs) per image (Table~\ref{tab:main}). All pretrained weights
were taken from the timm library \citep{wightman2019timm}, and all layers were fine-tuned.

We also trained two models from scratch, that is, without pretrained weights. They represent the compact custom CNNs that are
often proposed for waste classification \citep{zheng2023focusrcnet}. The first is a compact custom CNN, which we call V0. It has 0.38 million parameters and needs 165 million MACs. It combines residual blocks, a multi-scale
depthwise convolution module, channel and spatial attention, and a dual-pooling head. It was implemented in PyTorch. The second is MobileNetV3-L trained from scratch. Because it shares the architecture of the pretrained
MobileNetV3-L, this pair separates the effect of pretraining from the effect of the design.

\subsection{Training}\label{sec:training}
All models were trained with one shared recipe, so that only the model changed between runs (Table~S1). Images were rotated
according to their EXIF tag and resized so that the shorter side was 256 pixels. Training used random $224 \times 224$ crops,
flips, small rotations and small colour changes. With a probability of 0.5, a training batch was also mixed with MixUp
\citep{zhang2018mixup} or CutMix \citep{yun2019cutmix}, which blend two images and their labels. The models were optimised
with AdamW \citep{loshchilov2019adamw} and a batch size of 32. The learning rate rose during five warm-up epochs and then
followed a cosine decay \citep{loshchilov2017sgdr}. We evaluated an exponential moving average of the weights after the last
epoch \citep{polyak1992averaging}. No early stopping, test-time augmentation or class weighting was used. Training ran in
PyTorch \citep{paszke2019pytorch} with timm on an NVIDIA Tesla T4 graphics processing unit (GPU).

Each pretrained model received its own learning rate. We tried $10^{-3}$, $3 \times 10^{-4}$ and $10^{-4}$ on the tuning split.
We then chose the value with the highest mean validation macro-averaged F1 score (macro-F1, Section~\ref{sec:metrics}) over
epochs 16 to 40 (Table~S2). This gives every model a fair
setting. All pretrained models were trained for 40 epochs, because their validation curves reached a plateau between epochs 20
and 40 (Fig.~S4). ConvNeXt-T and DeiT-S declined afterwards. ConvNeXt-T, for example, fell from
about 93\% to 89\% by epoch 80, which suggests memorisation of the training objects. The two models trained from scratch used a
learning rate of $10^{-3}$ and a longer schedule of 150 epochs.

\subsection{Metrics and statistical analysis}\label{sec:metrics}
Macro-F1 on unseen objects is the main metric. It is the mean of the F1 scores of the 23 classes, so every class counts
equally. We also report accuracy and balanced accuracy. Each metric is given as the mean and standard deviation over
the five test folds. We also pooled the out-of-fold predictions, in which each of the 4,160 images is predicted once by a
model that never saw its object. For the pooled macro-F1, we computed a 95\% confidence interval by
bootstrap \citep{efron1979bootstrap} with 2,000 resamples. We resampled the 172 objects rather than single images, because
images of the same object are not independent \citep{field2007clustered}. To compare models, we applied the McNemar test with
continuity correction \citep{mcnemar1947note,dietterich1998approximate} to the pooled predictions of each model and of the
best model. The $p$-values of the 12 comparisons were adjusted with the Holm method \citep{holm1979simple}. For cost, we report the number of parameters and the MACs per image.

\subsection{Additional analyses}\label{sec:analyses}
Five further analyses looked beyond overall accuracy. First, we counted errors between the three groups, in particular
medical images predicted as non-medical. Second, we computed the accuracy of each object. Third, we tested ConvNeXt-T and
EfficientNet-B0 for image-quality shortcuts. The test images were re-encoded as JPEG files at quality 30, or reduced to 96
pixels and enlarged again, and accuracy was compared between phones. Fourth, we computed Grad-CAM maps on fold 1
\citep{selvaraju2017gradcam}. Fifth, four models trained on all 4,160 images were tested on 1,390 BDWaste images from Dhaka
\citep{rahman2024bdwaste}. These images belong to the 11 classes shared by the two datasets. We report accuracy over all 23
outputs and accuracy restricted to the shared classes.

\section{Results}\label{sec:results}

\subsection{Classification of unseen objects}\label{sec:res_main}
We first evaluated all models on unseen objects with object-disjoint cross-validation (Table~\ref{tab:main}). ConvNeXt-T
reached the highest pooled macro-F1 of 98.82\%, with a 95\% confidence interval of 94.5 to 99.4\%. The interval is wide at its
lower end because the bootstrap resamples only 172 objects. All 11 pretrained models scored between 96.03 and 98.82\%, and
MobileNetV4-S had the lowest score among them.

Six models could not be separated statistically. ResNet-50, MobileViTv2, DenseNet-121, DeiT-S and EfficientNet-B0 did not
differ significantly from ConvNeXt-T (Holm-corrected $p$ between 0.059 and 0.272). The other five pretrained models and both
models trained from scratch did differ ($p < 0.001$). 

\begin{figure}[!htbp]
\centering
\includegraphics[width=\figwidth]{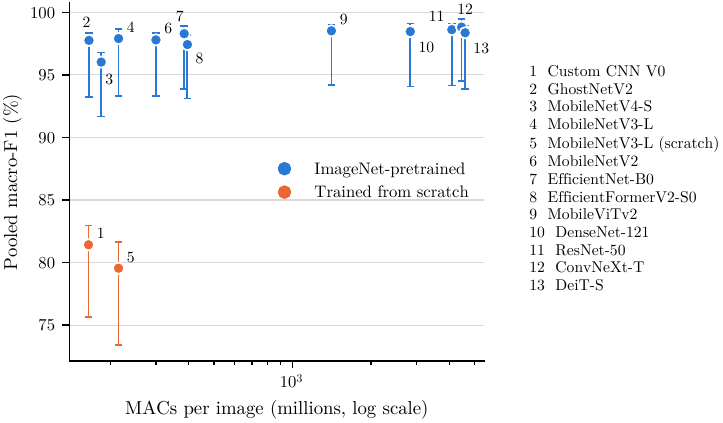}
\caption{Pooled macro-F1 on unseen objects against computational cost (MACs per image, log scale). Error bars show 95\% confidence
intervals (bootstrap over objects). Blue points show ImageNet-pretrained models, and orange points show models trained from scratch.
Numbers identify the models listed on the right.}
\label{fig:macs}
\end{figure}

Next, we related accuracy to computational cost (Fig.~\ref{fig:macs}). EfficientNet-B0 reached 98.29\% with 385 million MACs
and 4.04 million parameters. MobileViTv2 reached 98.52\% with 1,412 million MACs. ConvNeXt-T needed 4,455 million MACs and 27.84
million parameters for its 98.82\%. The statistically tied group therefore includes models that need 3 to 12 times less
computation than the best model.  The fifth fold was the hardest for every model. On this fold, the macro-F1 of the pretrained models
fell to between 89.0 and 93.8\%.

\begin{table}[!htbp]
\centering
\footnotesize
\caption{Classification of unseen objects (object-disjoint 5-fold cross-validation, 4,160 images). Macro-F1 is given as the mean $\pm$ SD over folds and pooled over all out-of-fold predictions with a 95\% confidence interval from an object-level bootstrap (2,000 resamples). $p$: McNemar test against the best model (ConvNeXt-T), Holm-corrected.}
\label{tab:main}
\begin{adjustbox}{max width=\textwidth}
\begin{tabular}{@{}lrrlccc@{}}
\toprule
Model & Params (M) & MACs (M) & LR & Macro-F1 (\%), fold mean & Pooled macro-F1 (\%) [95\% CI] & $p$ \\
\midrule
\multicolumn{7}{@{}l}{\textit{ImageNet-pretrained, fully fine-tuned (40 epochs)}} \\
ConvNeXt-T & 27.84 & 4,455 & 0.0003 & 97.99 $\pm$ 2.69 & \textbf{98.82} [94.5, 99.4] & -- \\
ResNet-50 & 23.56 & 4,087 & 0.0003 & 97.76 $\pm$ 2.75 & 98.62 [94.2, 99.1] & 0.27 \\
MobileViTv2 & 4.40 & 1,412 & 0.001 & 97.71 $\pm$ 2.40 & 98.52 [94.2, 99.0] & 0.27 \\
DenseNet-121 & 6.98 & 2,833 & 0.0001 & 97.65 $\pm$ 2.45 & 98.47 [94.1, 99.1] & 0.18 \\
DeiT-S & 21.67 & 4,599 & 0.0001 & 97.43 $\pm$ 2.84 & 98.37 [93.9, 99.0] & 0.06 \\
EfficientNet-B0 & 4.04 & 385 & 0.0003 & 97.43 $\pm$ 2.61 & 98.29 [93.9, 98.9] & 0.06 \\
MobileNetV3-L & 4.23 & 215 & 0.001 & 97.05 $\pm$ 2.79 & 97.91 [93.3, 98.7] & $<$0.001 \\
MobileNetV2 & 2.25 & 300 & 0.0003 & 97.07 $\pm$ 2.27 & 97.80 [93.3, 98.4] & $<$0.001 \\
GhostNetV2 & 4.91 & 166 & 0.0003 & 96.78 $\pm$ 3.41 & 97.76 [93.3, 98.3] & $<$0.001 \\
EfficientFormerV2-S0 & 3.25 & 396 & 0.001 & 96.60 $\pm$ 3.16 & 97.43 [93.1, 98.2] & $<$0.001 \\
MobileNetV4-S & 2.52 & 185 & 0.0001 & 95.11 $\pm$ 4.02 & 96.03 [91.7, 96.8] & $<$0.001 \\
\midrule
\multicolumn{7}{@{}l}{\textit{Trained from scratch (150 epochs)}} \\
Custom CNN (V0) & 0.38 & 165 & 0.001 & 79.92 $\pm$ 3.27 & 81.42 [75.6, 83.0] & $<$0.001 \\
MobileNetV3-L & 4.23 & 215 & 0.001 & 78.14 $\pm$ 4.61 & 79.54 [73.4, 81.6] & $<$0.001 \\
\bottomrule
\end{tabular}
\end{adjustbox}
\end{table}

\subsection{Random versus object-disjoint evaluation}\label{sec:res_leakage}
\begin{figure}[!htbp]
\centering
\includegraphics[width=\figwidth]{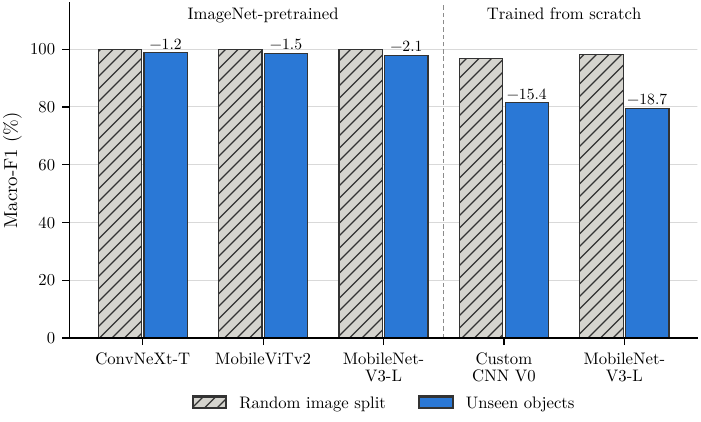}
\caption{Macro-F1 of the same models under a random image split (hatched grey bars) and on unseen objects (solid blue bars, object-disjoint cross-validation, pooled out-of-fold). Numbers above the blue bars give the drop in percentage points. The three models on the left are ImageNet-pretrained, and the two on the right were trained from scratch.}
\label{fig:leakage}
\end{figure}

We then repeated the evaluation with a random image-level split (Fig.~\ref{fig:leakage}, and Table~S3 in Online Resource 1). On this split,
ConvNeXt-T, MobileViTv2 and the pretrained MobileNetV3-L all reached a macro-F1 of 100.0\%. On unseen objects, the same models
reached 98.82, 98.52 and 97.91\%. The gap was therefore only 1.2 to 2.1 percentage points. However, the random split gave all
three models exactly the same score. It could not show which of them generalises better.

The gap was much larger for the models trained from scratch. The custom CNN V0 reached 96.87\% on the random split and 81.42\%
on unseen objects, a gap of 15.4 points. MobileNetV3-L trained from scratch fell from 98.21 to 79.54\%, a gap of 18.7 points. On
the random split, this model was only 1.8 points below its pretrained version. On unseen objects, it was 18.4 points below. The
random split thus made a weak model look almost as good as a strong one.

\subsection{Medical waste and group-level errors}\label{sec:res_medical}
Next, we examined the errors that matter most for sorting, which are those that cross a group boundary
(Table~\ref{tab:medical}). All pretrained models assigned 99.30 to 99.95\% of the images to the correct group. Their recall
for medical waste was between 99.10 and 100\%. At most 7 of the 781 medical images were predicted as non-medical, by
MobileNetV4-S. Four models missed none.

The models trained from scratch made many more medical errors. The custom CNN V0 reached a group accuracy of 94.95\% and a
medical recall of 89.12\%. It sent 85 medical images to a non-medical class and 52 non-medical images to a medical class.
MobileNetV3-L trained from scratch reached 93.94\% and 87.20\%, with 100 and 52 such errors. These two models therefore missed
about 11 to 13 of every 100 medical items. Syringes and medicine blister packs were among their weakest classes, with F1 scores
of 58.7 to 73.4\% (Table~S4).

\begin{table}[!htbp]
\centering
\footnotesize
\caption{Accuracy at the level of the three waste groups (food and plant, household non-food, medical) and errors involving medical waste (pooled out-of-fold predictions; 781 medical images).}
\label{tab:medical}
\begin{adjustbox}{max width=\textwidth}
\begin{tabular}{@{}lccccc@{}}
\toprule
Model & Group acc. (\%) & Medical recall (\%) & Medical precision (\%) & Medical $\rightarrow$ other & Other $\rightarrow$ medical \\
\midrule
ConvNeXt-T & 99.95 & 99.74 & 100.00 & 2 & 0 \\
ResNet-50 & 99.93 & 100.00 & 99.87 & 0 & 1 \\
MobileViTv2 & 99.93 & 100.00 & 100.00 & 0 & 0 \\
DenseNet-121 & 99.90 & 100.00 & 99.74 & 0 & 2 \\
DeiT-S & 99.78 & 99.36 & 99.87 & 5 & 1 \\
EfficientNet-B0 & 99.76 & 99.87 & 99.87 & 1 & 1 \\
MobileNetV3-L & 99.69 & 99.74 & 99.36 & 2 & 5 \\
MobileNetV2 & 99.62 & 99.23 & 99.61 & 6 & 3 \\
GhostNetV2 & 99.90 & 100.00 & 100.00 & 0 & 0 \\
EfficientFormerV2-S0 & 99.78 & 99.62 & 99.87 & 3 & 1 \\
MobileNetV4-S & 99.30 & 99.10 & 98.98 & 7 & 8 \\
\midrule
Custom CNN V0 (from scratch) & 94.95 & 89.12 & 93.05 & 85 & 52 \\
MobileNetV3-L (from scratch) & 93.94 & 87.20 & 92.91 & 100 & 52 \\
\bottomrule
\end{tabular}
\end{adjustbox}
\end{table}

\subsection{Difficult classes and objects}\label{sec:res_hard}
We then looked at which classes caused the remaining errors (Fig.~S5, Table~S4). For the pretrained models, the lowest F1
scores were found in four peel classes, namely potato, pointed gourd, cucumber and mango. ConvNeXt-T, for example, reached 91.7\%
for potato peel and 95.4 to 97.0\% for the other three. All its other classes scored between 97.6 and 100\%. These peels were
mostly confused with other food residues, such as another peel or peanut shells (Figs.~S6 and S7).

Object IDs showed that these errors were concentrated in a few objects (Table~S5). One potato peel object with seven images was
misclassified by all 11 pretrained models. One mango peel object with seven images was recognised in only 10.4\% of cases on
average, and one lemon object with six images in 45.5\%. A larger potato peel object with 48 images reached 80.1\%. Three of
the five hardest objects belong to the fifth fold, which explains why this fold was the hardest for every model. Such failures
stay hidden when results are reported only per class or per image.

\subsection{Effect of image quality and device}\label{sec:res_shortcut}
We next tested whether the models relied on image quality rather than on the waste item itself (Fig.~\ref{fig:transfer}b,
Table~S6). This check matters because all malta peel images are small, low-resolution files that a model could learn to detect. Strong JPEG compression lowered the
macro-F1 of ConvNeXt-T from 97.99 to 96.71\%, and resolution loss lowered it to 96.37\%. For EfficientNet-B0, the values fell from
97.43 to 96.06 and 94.71\%. The losses were therefore 1.3 to 2.7 points. If a model had used file quality as a cue, degraded
images of other classes would have been predicted as malta peel. This happened for no image with ConvNeXt-T and for four images
with EfficientNet-B0.

We also compared accuracy between phones within the same class (Table~S7). Thirteen classes were photographed with two phones.
In 11 of them, the accuracy of the three pretrained models tested differed by at most 3.6 points between phones. The exceptions
were mango peel and potato peel, with differences of 9.7 to 27.6 points. In both classes, the hardest objects of
Section~\ref{sec:res_hard} were all photographed with the Galaxy M31. The difference therefore reflects object difficulty rather
than the device.

\subsection{Cross-region test}\label{sec:res_bdwaste}
\begin{figure}[!htbp]
\centering
\includegraphics[width=\figwidth]{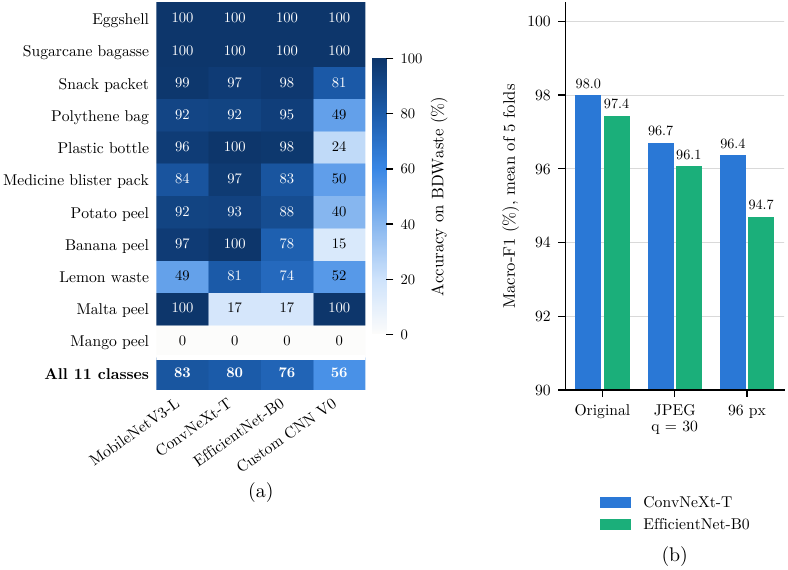}
\caption{(a) Accuracy on BDWaste images from Dhaka for the 11 classes shared with BanglaWaste-23 (models trained on all BanglaWaste-23 images). The last row gives the accuracy over all 1,390 images. (b) Macro-F1 (mean of five folds) on the
object-disjoint test folds for the original images, after strong JPEG compression (quality 30) and after reducing the
resolution to 96 pixels.}
\label{fig:transfer}
\end{figure}

Finally, we tested the models on images from another region (Fig.~\ref{fig:transfer}a, Table~S8). On the 1,390 BDWaste images
from Dhaka, with predictions restricted to the 11 shared classes, MobileNetV3-L reached the highest accuracy of 83.0\%. ConvNeXt-T
reached 80.1\%, EfficientNet-B0 76.0\% and the custom CNN V0 55.8\%.  For the pretrained models, the restricted accuracy was 15 to 22 points below their accuracy on unseen objects of
BanglaWaste-23.

The drop was concentrated in a few classes. No model recognised any of the BDWaste mango peel images. For malta peel, ConvNeXt-T
and EfficientNet-B0 were correct for only 17\% of the images. MobileNetV3-L and V0, in contrast, were correct for all of them in
the restricted setting.  One possible reason for the
failures is that mango and malta peels look different in the two datasets. We did not test this explanation directly.

\subsection{Visual explanations}\label{sec:res_gradcam}
Grad-CAM maps showed where the models looked when they made a prediction (Figs.~S8 to S15). The pretrained models focused
on the waste item itself, such as the barrel of a syringe, a blister pack or an eggshell. The two models trained from scratch
produced patchy maps. These maps often covered the background, for example the white area above a mango leaf or the ring around
an orange peel. The maps come from one fold and are qualitative. They suggest that the models trained from scratch rely partly on
background cues. They do not prove it.

\section{Discussion}\label{sec:discussion}
Automatic sorting is only useful if it works on items that the system has never seen. Waste classification studies often
report accuracies close to 100\%. Most of them, however, test on images of the same items that were used for training. It was
therefore unclear how much of this accuracy remains for new items. It was also unclear how safe the errors are for medical
waste and how well the models work in another region. Our results lead to three main findings.

\subsection{What random splits hide}\label{sec:disc_split}
First, a random image-level split does not measure what a sorting system needs. In our random split, 155 of the 172 objects
had images in both the training and the test set. The test therefore asked a model to recognise an item that it had already
seen from another angle. Every pretrained model passed this test perfectly. As a result, the random split could not rank the
models. Object-disjoint evaluation, in contrast, separated them clearly.

The random split also hid the failure of models trained from scratch. The same MobileNetV3-L architecture was nearly perfect on
the random split and lost about 18 points on unseen objects. The cause is therefore training from scratch on only 5 to 10
objects per class, not one particular design. The custom CNN V0 showed the same pattern, with 96.9\% on the random split and 81.4\% on unseen objects. Compact custom CNNs trained from scratch are common in the waste
literature, and most studies evaluate on random splits (Section~\ref{sec:rw_leakage}). Their reported accuracies should
therefore be read as an upper bound for unseen items. Such results remain valid for items that were seen during training.
They answer a different and easier question. 

\subsection{Implications for waste-sorting systems}\label{sec:disc_practice}
Second, medical waste was handled safely only by pretrained models. The models trained from scratch missed about 11 to 13 of
every 100 medical items. In a sorting system, these items would be treated as ordinary household waste. Health-care waste is a
small part of all waste. A part of it, however, is hazardous \citep{who2024healthcarewaste}. The pretrained models missed at
most 7 of 781 medical images, and four of them missed none. For medical waste, the number of medical items sent to other
classes is therefore a more useful figure than overall accuracy.

Third, small pretrained models were accurate enough on unseen objects. They did not, however, transfer well to another region.
EfficientNet-B0 was statistically tied with ConvNeXt-T at about one-twelfth of the computation. This makes it a reasonable
choice for low-cost sorting hardware. We did not measure speed on such hardware, so this point rests on the number of
operations only. On BDWaste images from Dhaka, the accuracy of the pretrained models fell by 15 to 22 points, and no model
recognised mango peel. A model trained in one city therefore needs images from the new location, and probably some local
fine-tuning, before it is used there.

These findings lead to three suggestions for future waste datasets and studies. Results should be reported on unseen objects
with confidence intervals. Object IDs should be released together with the images. Where possible, models should also be tested
on data from another site.

\subsection{Pretraining and architecture}\label{sec:disc_custom}
In a pilot on the validation objects of the
tuning split (Table~S9 in Online Resource 1), the custom CNN V0 trained from scratch reached a macro-F1 of 70.8\%. Knowledge distillation from
ConvNeXt-T \citep{hinton2015distilling} raised it to 74.2\%.  The pretrained
MobileNetV3-L reached 90.2\% on the same objects. Adding our multi-scale and attention modules to this backbone gave
87.3\% without distillation and 91.1\% with it, within about one point of the plain backbone. These are single validation runs and should be read with caution. They still show that, with 5 to 10 objects per
class, pretraining mattered far more than the architecture. Larger and more varied datasets may be needed before changes to the architecture show a clear benefit.

\subsection{Limitations}\label{sec:limitations}
We acknowledge that this study has a few limitations. The images come from two districts and one hospital. Each class has 5 to
10 objects, so the confidence intervals are fairly wide. Object IDs were assigned by visual grouping and a manual review, so a
few views of the same item may still fall into different objects. Most items were photographed one at a time and indoors.
Sorting lines, in contrast, handle mixed and overlapping waste. Finally, we report computational cost but not speed on a
device. These points are addressed in the future work outlined in Section~\ref{sec:conclusion}.

\section{Conclusion}\label{sec:conclusion}
This study introduced BanglaWaste-23, an image dataset of household and hospital waste from northern Bangladesh in which every
image is linked to the physical object that it shows. We used the object IDs to build an object-disjoint benchmark and
evaluated 11 pretrained models and two models trained from scratch on objects that they had never seen.

Pretrained models recognised unseen objects almost perfectly, and small models such as EfficientNet-B0 were as accurate as much
larger ones. They also kept medical items apart from household waste with very few errors. A random image-level split, in
contrast, gave every pretrained model a perfect score and could not rank them. It also hid a loss of 15 to 19 percentage points
for models trained from scratch, which missed far more medical items. On images from Dhaka, all models lost accuracy, and some
classes were not recognised at all.

These results come from two districts, one hospital and single items photographed mostly indoors. They do not yet show how the
models behave on mixed waste in a real sorting line. Even so, they lead to a clear lesson for waste-sorting research. Accuracy
should be measured on unseen objects, and results from random image-level splits should be read as an upper bound. The dataset,
its object IDs and our split files are released to support this practice.

This work is a necessary first step towards reliable image-based sorting of household and medical waste. Future work will add
more regions and hospitals, extend the data to mixed-waste scenes and object detection, and measure speed on low-cost devices.

\section*{Author contributions}
Nusrat Jahan: Investigation, Data curation, Software. Mst. Eshrat Jahan Esha: Investigation, Data curation, Formal analysis.
Masuma Sadia: Data curation. Saad Ahmed: Data curation, Methodology, Software, Formal analysis,
Supervision, Writing -- original draft, Writing -- review \& editing. Md Zahim Hassan: Supervision, Writing -- review \&
editing. S M Jahangir Alam: Supervision, Writing -- review \& editing.

\section*{Declaration of competing interest}
The authors declare that they have no known competing financial interests or personal relationships that could have
appeared to influence the work reported in this paper.

\section*{Funding}
This research did not receive any specific grant from funding agencies in the public, commercial, or not-for-profit sectors.

\section*{Ethics statement}
The images show discarded objects only. No people, patient names or other personally identifiable information appear in the
dataset. Medical waste items were collected at Hasan General Hospital, Thakurgaon, with the permission of the hospital's
Managing Director. Team members wore gloves and face masks while handling these items, and the items were returned to the
hospital after photography.

\section*{Data availability}
The BanglaWaste-23 dataset is available on Zenodo at \url{https://doi.org/10.5281/zenodo.23694940} under the CC BY-NC 4.0
licence \citep{jahan2026banglawaste23}. The record is restricted during peer review, and reviewers can request access through
Zenodo. The record also contains the object-disjoint split files used in this paper. The training and analysis code is available at
\url{https://github.com/saadbaust/BanglaWaste-23}.

\section*{Declaration of generative AI and AI-assisted technologies in the manuscript preparation process}
During the preparation of this work the authors used Claude (Anthropic) to support the literature search, to write the
training and analysis code, and to draft and edit the manuscript text. After using this tool, the authors reviewed and edited
the content as needed and take full responsibility for the content of the published article.

\section*{Acknowledgements}
The authors thank the Managing Director of Hasan General Hospital, Thakurgaon, for permission to collect medical waste items,
and the hospital nurse who verified the labels of the medical classes.

\section*{Supplementary information}
Supplementary material associated with this article (Figs.~S1 to S15 and Tables~S1 to S9) is provided in Online Resource 1, which follows the references.

\bibliographystyle{unsrtnat}
\bibliography{references}

\clearpage
\includepdf[pages=-]{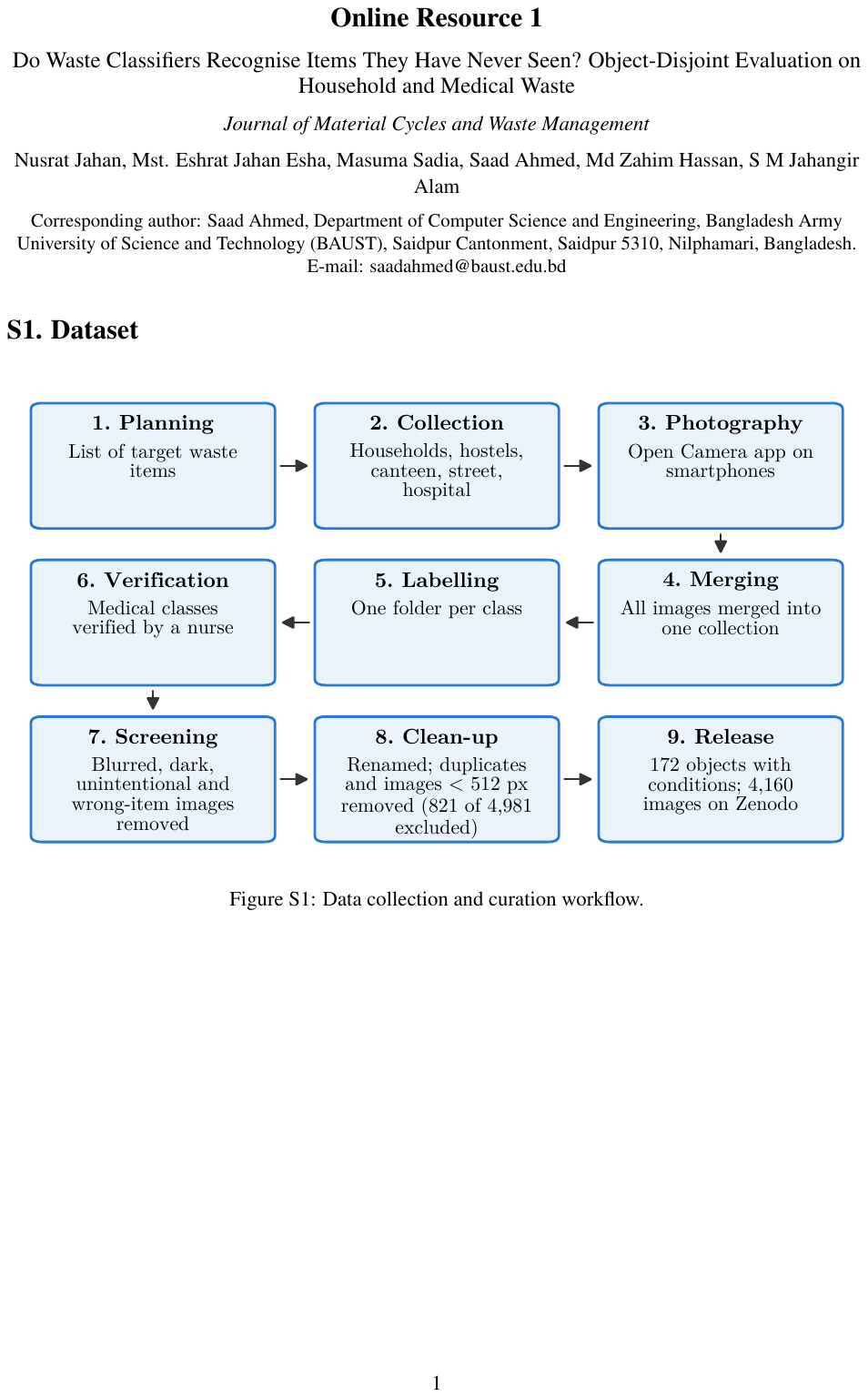}
\end{document}